\documentstyle{nemlap} 
\input{ipamacs}
\input{epsf}
\begin{document}
\title{Morphological Analysis as Classification: \\
an Inductive-Learning Approach} 
\author{Antal van den Bosch$^{(i)}$, 
        Walter Daelemans$^{(ii)}$,
	Ton Weijters$^{(i)}$}
\institute{
$^{(i)}$ Dept. of Computer Science / {\sc matriks} \\
University of Limburg $|$ Maastricht \\
PO Box 616, 6200 MD Maastricht, THE NETHERLANDS 
\ \\[0.2cm]
$^{(ii)}$ Computational Linguistics \\
Tilburg University \\
PO Box 90153, 5000 LE Tilburg, THE NETHERLANDS}
\maketitle

\begin{abstract}
Morphological analysis is an important subtask in text-to-speech
conversion, hyphenation, and other language engineering tasks. The
traditional approach to performing morphological analysis is to
combine a morpheme lexicon, sets of (linguistic) rules, and heuristics
to find a most probable analysis. In contrast we present an inductive
learning approach in which morphological analysis is reformulated as a
segmentation task. We report on a number of experiments
in which five inductive learning algorithms are applied to three
variations of the task of morphological analysis.  Results show (i)
that the generalisation performance of the algorithms is good, and
(ii) that the {\sl lazy learning}\/ algorithm {\sc ib1-ig} performs
best on all three tasks. We conclude that lazy learning of
morphological analysis as a classification task is indeed a viable
approach; moreover, it has the strong advantages over the traditional
approach of avoiding the knowledge-acquisition bottleneck, being fast
and deterministic in learning and processing, and being
language-independent. 
\end{abstract}

\section{Introduction}
\label{intro}

Morphological analysis is often deemed to be an important, if not
essential subtask in linguistic modular systems for text-to-speech
processing \cite{allen-mitalk} and hyphenation
\cite{daelemans-chyp}. In text-to-speech processing, it serves to
prevent the wrong application of grapheme-phoneme conversion rules
across morpheme boundaries (e.g., preventing {\sf carelessly} from
being pronounced as /k\schwa 'r\niepsilon l\schwa slai/). In
hyphenation (in British English and to a lesser degree in Dutch), it
guides the placement of hyphens at certain morpheme boundaries
(e.g., preventing {\sf looking} from being hyphenated as {\sf
loo-king}). Morphological analysis also plays a crucial role in
applications such as part-of-speech tagging (assigning the correct
morpho-syntactic category to words in context), for obtaining a
reasonable analysis of words not present in the lexicon.

The traditional approach to performing morphological analyses
presupposes the availability of a morpheme lexicon, spelling rules,
morphological rules, and heuristics to prioritise possible analyses of
a word according to their plausibility (e.g., see the {\sc decomp}
module in the {\sc mit}talk system \cite{allen-mitalk}). In
contrast, the approach described in this paper presupposes a
morphologically analysed corpus of words (rather than a corpus of
morphemes), and an inductive learning algorithm trained to
segment spelling words into morphemes. This segmentation task is
reformulated as a simple classification task.

In this paper, we will first outline what we mean by rephrasing a
linguistic problem as a classification task, and we will introduce
five inductive-learning algorithms which are applied to this
task. Then, in section~\ref{morpanal}, we give an
overview of the traditional approach to morphological analysis and
introduce our alternative reformulation. In section~\ref{exps} we
present and analyse the results of the application of the learning
algorithms to this classification task. We conclude this paper with a
summary of the obtained results and a discussion of the differences
between the traditional approach to morphological analysis and a
inductive-learning approach, in section~\ref{conclusions}.

\subsection{Reformulating Linguistic Problems as Classification Tasks}

Most linguistic problems can be seen as context-sensitive mappings
from one representation to another (e.g., from text to speech; from a
sequence of spelling words to a parse tree; from a parse tree to
logical form, from source language to target language, etc.).  The
typical traditional approach to language engineering problems is to
build a description of the general rules governing these mappings,
describe additional subregularities, and list the remaining exceptions
to the rules and subregularities. The acquisition of this knowledge is
labour-intensive and costly. In contrast to this hand-crafting
approach, an inductive machine-learning method approaches a linguistic
problem in a data-oriented way, i.e., it automatically gathers the
knowledge needed for solving the problem by considering 
instances of the problem. By `instance' we mean a data structure
containing an input and its associated `solution'; its classification.
The knowledge implicitly present in the collection of instances is
used to classify new instances of the same problem.

Most linguistic tasks can be described as classification tasks, i.e.,
given a description of an input in terms of a number of 
feature-values, a classification of the input is performed. Two types of
classification tasks can be discerned \cite{daelemans-mem}:

\begin{itemize}
\item
{\bf Identification}: given a set of possible classifications and an
input of feature values, determine the correct classification for
this input. For example, given a letter surrounded by a number of
neighbours (e.g., {\sf a} in {\sf have}), determine the phonemic
transcription of that letter.
\item
{\bf Segmentation}: given a set of possible boundary classes and an
input consisting of a focus position in its immediate context,
determine whether a boundary is associated with the focus position,
and if so, which one. For example, determine if the {\sf b} in {\sf
table} marks the boundary of a syllable.
\end{itemize}

Once a task is reformulated as a classification task, it can be
learned by an inductive-learning algorithm. Differences exist in the
ways inductive-learning algorithms extract knowledge 
from the available instances. In {\sl lazy learning} (such as 
memory-based learning \cite{stanfill-mbr,daelemans-mem}), there is no
abstraction of 
higher-level data structures such as rules or decision trees at learning
time; learning consists of simply storing the instances in memory.  A
new instance of the same problem is solved by retrieving those
instances from memory that match the new instance best (according to a
similarity metric), and by extrapolating from the solutions of these
`nearest neighbours'.  The {\sl memory-based learning approach}
therefore does not distinguish between regularities and individual
exceptions; rule-like behaviour is the result of the interaction
between the memory contents and the similarity metric used. In {\sl eager
learning} approaches (such as C4.5 or connectionist learning), abstract
data structures (matrices of connection weights in connectionist
networks, decision trees in C4.5) are extracted from the learning
material during learning.

In previous research we have demonstrated the application of the
memory-based (lazy) learning approach to several linguistic problems,
e.g., segmentation as in hyphenation and syllabification
\cite{daelemans-hyphen,vandenbosch-benelearn95}, and identification as
in grapheme-phoneme conversion
\cite{weijters-nettalk,vandenbosch-grafon,daelemans-independent}, and
stress assignment \cite{daelemans-cl}.  In most cases, the memory-based
(lazy) approach outdid the more eager inductive algorithms. We
believe that in a `noisy' domain such as natural language, abstracting
from the training instances is a bad idea because any one instance
(however `exceptional' from the point of view of the learning
algorithm) can potentially be a model for new instances.

In this paper, we will demonstrate that the segmentation approach of
memory-based learning is also applicable to morphological parsing. We
will compare the approach to alternative inductive machine-learning
algorithms. First, we provide a brief summary of the
inductive-learning algorithms used in the experiments reported in
this paper.

\subsection{Algorithms and Methods for Inductive Learning} 

Inductive learning in its most straightforward form is exhibited by
memory-based {\sl lazy learning}\/ algorithms such as {\sc ib1}
\cite{aha-ibl} and variations (e.g., {\sc ib1-ig}
\cite{daelemans-hyphen,daelemans-air}), in which all instances are 
fully stored in 
memory, and in which classification involves a pass along all stored
instances. To optimise memory lookup and minimise memory usage, more
eager learning algorithms are available that compress the memory in
such a way that most relevant knowledge is retained and stored in
a quickly accessible form, and redundant 
knowledge is removed. Examples of such algorithms are the
decision-tree algorithms {\sc igtree} \cite{daelemans-air} and {\sc
c4.5} \cite{quinlan-c45}. Another popular inductive algorithm is the
connectionist Back-propagation ({\sc bp}) \cite{rumelhart-bp} learning
algorithm. We provide a summary of the basic functions of these
learning algorithms.

\begin{enumerate}
\item
{\sc ib1} \cite{aha-ibl} constructs a data base of instances (the {\sl
instance base}) during learning. An instance consists of a
fixed-length vector of $n$ feature-value pairs, and an information
field containing the classification(s) of that particular
feature-value vector. When the feature-value vector is associated to
more than one classification (i.e., when its classification is
ambiguous), the occurrences of the different classifications in the
learning material are counted and stored with the instance. After the
instance base is built, new instances are classified by {\sc ib1} by
matching them to all instances in the instance base, and calculating
with each match the {\sl distance}\/ between the new instance $X$ and
the memory instance $Y$, $\Delta(X,Y)$, using the function in
equation~\ref{simfunc}.

\begin{equation}
\Delta(X,Y)=\sum_{i=1}^{n}W(f{_i})\delta(x_{i},y_{i})
\label{simfunc}
\end{equation}

where $W(f{i})$ is the weight of the
$i$th feature (in {\sc ib1}, this weight is equal for all features),
and $\delta(x_{i},y_{i})$ is the distance 
between the values of the $i$th feature in instances $X$ and $Y$.
 
When the values of the instance features are symbolic, as with our
linguistic tasks, a simple distance function for
$\delta(x_{i},y_{i})$ is used (equation \ref{simpdist}).
 
\begin{equation}
\delta(x_{i},y_{i}) = 0 \; if \; x_{i}=y_{i}, \; else \; 1
\label{simpdist}
\end{equation}

The (most frequently occurring) classification of the memory instance
$Y$ with the smallest $\Delta(X,Y)$ is then taken as the classification
of $X$.

\item
{\sc ib1-ig} \cite{daelemans-hyphen,daelemans-air} differs from {\sc
ib1} in the 
weighting function $W(f{_i})$ (cf. equation~\ref{simfunc}). This 
function computes for each feature, over the full instance base, its
{\sl information gain}, a function from information theory that is
also used in {\sc id3} \cite{quinlan-id3} and {\sc c4.5}
\cite{quinlan-c45} (for more details, cf. Daelemans and Van den 
Bosch \cite{daelemans-hyphen}). In short, the information gain of a feature expresses
its relative importance compared to the other features in performing
the mapping from input to classification. This weighting function gives right to
the fact that for some tasks, some features are far more important
than other features. When information gain is used as the weighting
function in the similarity function (equation~\ref{simfunc}), instances
that match on an important feature are regarded as more alike than
instances that match on an unimportant feature.

\item
{\sc igtree} \cite{daelemans-air} compresses an instance base into a
decision tree. Instances are stored in the tree as paths of connected
nodes, and leaves containing classification information. Nodes are
connected via arcs denoting feature values. Information gain is used
in {\sc igtree} to determine the order in which instance feature
values are added as arcs to the trie. The reasoning behind this
compression is that when the computation of information gain points to
one feature clearly being the most important in classification, search
can be restricted to matching a test instance to those memory
instances that have the same feature value as the test instance at
that feature. Instead of indexing all memory instances only once on
this feature, the instance memory can then be optimised further by
examining the second most important feature, followed by the third
most important feature, etc.  A considerable compression is obtained
as similar instances share partial paths.  The trie structure is
compressed even more by restricting the paths to those input feature
values that disambiguate the classification from all other instances
in the training material. The idea is that it is not necessary to
fully store an instance as a path when only a few feature values of
the instance make the instance classification unique. In applications
to linguistic tasks, {\sc igtree} is shown to obtain compression
factors of 90\% or more as compared to {\sc ib1}/{\sc ib1-ig}
\cite{vandenbosch-grafon,daelemans-independent}.

{\sc igtree} also stores with each non-terminal node 
information concerning the {\sl most probable} or {\sl default} 
classification given the path thus far, according to the
classification bookkeeping
information maintained by the trie construction algorithm. This 
extra information is essential when processing
new instances. 
Processing a new instance involves traversing the trie (i.e.,
matching all feature-values of the test instance with arcs in
the order of the overall feature information gain), and
either retrieving a classification when a leaf is reached (i.e., an exact
match was found), or using the default classification on the last
matching non-terminal node if an exact match fails.
For more details on {\sc igtree}, see Daelemans et al. \cite{daelemans-air}.

\item
{\sc c4.5} \cite{quinlan-c45} is a well-known decision-tree algorithm
which basically uses the same type of strategy as {\sc igtree} to
compress an instance base into a compact tree. To this purpose,
standard {\sc c4.5} also uses information gain, or {\sl gain ratio}\/
\cite{quinlan-c45} to select the most important feature in tree
building; however, in
contrast to {\sc igtree}, {\sc c4.5} recomputes this function for each
node in the tree. Another difference with {\sc igtree} is that {\sc
c4.5} implements a pruning stage, in which parts of the tree are
removed as they are estimated to contribute to instance classification
below a certain utility threshold.

\item
{\sc bp} \cite{rumelhart-bp} is an artificial-neural-network learning
rule, which operates on multi-layer feed-forward networks ({\sc
mfn}s). In these networks, feature-values of instances are encoded as
activation patterns in the input layer, and the network is trained to
produce an activation pattern at the output layer representing the
desired classification. In contrast to the previously described
algorithms, {\sc bp} does not accumulate its knowledge by literally
storing (parts of) instances in memory or by constructing a decision
tree on the basis of them. Rather, {\sc bp} tunes the connections
between units in the input layer and the hidden layer, and between
units of the hidden layer and the output layer, during a training
phase in which all training instances are presented several times to
the network.
The {\sc bp} learning algorithm, which is a gradient descent
algorithm, attempts to set the connections between the layers with
increasing subtlety, aiming at minimisation of the error on the
training material. After training, the units at the hidden
layer encode an intermediary representation that captures (in an often
opaque way) some essential information from both the input (the
feature-values) and the output (the desired
classification). These representations are non-symbolic, and do not
lend themselves easily for inspection, in contrast to the previously
described symbolic algorithms.
\end{enumerate}

When one plans to apply learning algorithms to classification tasks,
it is important to establish a method for interpreting the results
from such experiments beforehand. In our experiments, we are primarily
interested in the {\sl generalisation accuracy}\/ of trained models,
i.e., the ability of these models to use their accumulated knowledge
to classify new instances that were not in the training material. A
method that gives a good estimate of the generalisation performance of
an algorithm on a given instance base, is {\sl 10-fold
cross-validation} \cite{weiss-10foldcv}. This method generates on the
basis of an instance base $10$ partitionings into a training set
(90\%) and a test set (10\%), resulting in $10$ experiments and $10$
results per learning algorithm and instance base. Significance tests
such as one-tailed t-tests can be applied to the outcomes of 10-fold
cross-validation experiments with several learning algorithms trained
on the same data.

\section{Morphological Analysis}
\label{morpanal}

\subsection{Traditional Approaches}

The traditional approach to morphological analysis basically
presupposes three components: (i) a morpheme lexicon, (ii) a set of
spelling rules and morphological rules to discover possible analyses
of morphologically complex words, and (iii) prioritising
heuristics to choose the most probable analysis from sets of possible
analyses. We briefly illustrate the functioning of this type of
analysis by taking {\sc decomp}'s processing as an example, and the word {\sf
scarcity} as the example word \cite{allen-mitalk}:

\begin{enumerate}
\item
In a morpheme lexicon covering the English language, a first analysis
divides {\sf scarcity} 
into {\sf scar} and {\sf city}.
\item
The analysis {\sf scar$|$city} is validated by a finite-state
automaton covering the possible sequences of morphemes
in English words; furthermore, an analysis-cost heuristic assigns an
integer-valued cost to the combination of the two noun stems.
\item
Using spelling rules for letter deletion in inflection and compounding
in English, the system suspects that the analysis {\sf scarce$|$ity}
is also possible, as {\sf ity} may have deleted the {\sf e} of {\sf
scarce}. This analysis, which is validated by the morpheme-sequence
finite-state automaton, yields a lower cost than {\sf scar$|$city},
as the analysis-cost heuristic assigns a lower value to a derivational
affix than to a second stem.
\item
As no further spelling-change rules can be applied to the analysis
with the lowest cost, {\sf scarce$|$ity}, the process ends by
producing this analysis. 
\end{enumerate}

It is argued in Allen et al. \cite{allen-mitalk} that a morpheme lexicon
containing 10,000 morphemes is effective in a text-to-speech
system. Neologisms, a problem for purely lexicon-based approaches,
seldomly contain new morphemes. The morpheme-sequence finite-state
automaton, the spelling rules, and the analysis-cost heuristic are in
principle not very complex in terms of processing. They demand,
however, a considerable amount of knowledge acquisition and
fine-tuning. Another serious problem with these analysis components is
that the number of analyses of morphologically complex words may
become very much larger (near exponential in the number of morphemes)
for longer words.

Morphological analysis on a probabilistic basis, using only a morpheme
lexicon, an analyses generator, and a probabilistic function to
determine the analysis with the highest probability
\cite{heemskerk-morpa} does not suffer from the disadvantageous
knowledge acquisition and fine-tuning phase, but is nevertheless also
confronted with an explosion of the number of generated analyses.

\subsection{Inductive-Learning Approach}

In contrast to this decomposition into three components, we
reformulate the task of morphological analysis as a one-pass
segmentation task, in which an input (a sequence of letters with a
focus position) is to be classified as marking a morpheme boundary at
that focus position. This classification approach demands that the
number of input features be fixed, hence we cannot use whole words as
input. Instead, we convert a word into fixed-sized instances of which
the middle letter is mapped to a class denoting a morpheme boundary
decision. To generate fixed-sized instances, we adopt the windowing
scheme proposed by Sejnowski and Rosenberg \cite{sejnowski-nettalk}
which generates 
fixed-sized snapshots of words.

\begin{figure}
\centerline{
        \epsfxsize=10cm
        \epsfbox{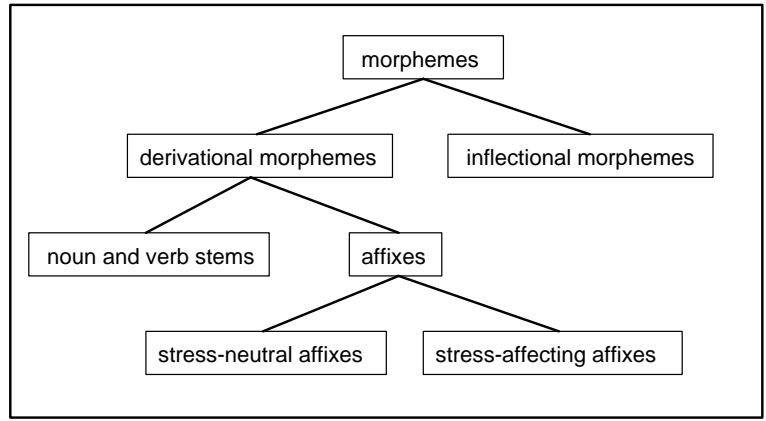} }
\center
\vspace*{-0.5cm}
\caption{Family tree of English morphemes.}
\label{morph-tree}
\end{figure}

In its most basic form, the classification of each instance denotes whether
the focus letter of the instance maps to a morpheme boundary (`{\sc
yes}', or `1') or 
not (`{\sc no}', or `0'). However, distinguishing between only 
`1' and `0' does not take into account that
morphological theory generally distinguishes between several types of
morphemes. For the case of English, a family tree of morphemes would
for example be the one displayed in Figure~\ref{morph-tree}.

Distinguishing between, for example, stress-neutral and
stress-affecting affixes would be directly helpful as input knowledge
for performing the stress-assignment task in a
text-to-speech system. However, distinguishing
between types of morphemes according to this theory also
introduces a certain amount of pre-wired linguistic knowledge. With
this in mind we extended the task of morphological analysis into three
different tasks, with increasing implicit linguistic knowledge encoded
in the classes: 
 
\begin{description}
\item[task {{\sc m1}}] -
decide whether the focus letter marks the beginning of
\begin{itemize}
\item a morpheme: map to class `1',
\item no morpheme: class `0'.
\end{itemize}
\item[task {{\sc m2}}] -
decide whether the focus letter marks the beginning of
\begin{itemize}
\item a derivational morpheme: class `d',
\item an inflectional morpheme: class `i',
\item no morpheme: class `0'.
\end{itemize}
\item[task {{\sc m3}}] -
decide whether the focus letter marks the beginning of
\begin{itemize}
\item a noun or verb stem: class `s',
\item a stress-neutral affix: class `1',
\item a stress-affecting affix: class `2',
\item an inflectional morpheme: class `i',
\item no morpheme: `0'.
\end{itemize}
\end{description}
 
Applying the windowing method to the example word {\sf
abnormalities} leads
to the instances displayed in Table~\ref{morph-window-ex}, listing for each
of the three tasks their appropriate classifications. The
morphological analysis of the full word is simply the concatenation of the
instance classifications, in which all classifications other than `0'
mark morpheme boundaries.
 
\begin{table}
\begin{center}
\begin{tabular}{|c|ccc|c|ccc|c|c|c|}
\hline
{\sc instance} & \multicolumn{3}{|c|}{\sc left} & {\sc focus} & 
\multicolumn{3}{|c|}{\sc right} &
\multicolumn{3}{|c|}{\sc task} \\
{\sc number} & \multicolumn{3}{|c|}{\sc context} & {\sc letter} & 
\multicolumn{3}{|c|}{\sc context}
& {\sc m1} & {\sc m2} & {\sc m3} \\
\hline
\hline
1  & {\sf \_} & {\sf \_} & {\sf \_} & {\sf a} & {\sf b} & {\sf n} & {\sf o} & 1 & d & 1 \\
\hline
2  & {\sf \_} & {\sf \_} & {\sf a} & {\sf b} & {\sf n} & {\sf o} & {\sf r} & 0 & 0 & 0 \\
\hline
3  & {\sf \_} & {\sf a} & {\sf b} & {\sf n} & {\sf o} & {\sf r} & {\sf m} & 1 & d & s \\
\hline
4  & {\sf a} & {\sf b} & {\sf n} & {\sf o} & {\sf r} & {\sf m} & {\sf a} & 0 & 0 & 0 \\
\hline
5  & {\sf b} & {\sf n} & {\sf o} & {\sf r} & {\sf m} & {\sf a} & {\sf l} & 0 & 0 & 0 \\
\hline
6  & {\sf n} & {\sf o} & {\sf r} & {\sf m} & {\sf a} & {\sf l} & {\sf i} & 0 & 0 & 0 \\
\hline
7  & {\sf o} & {\sf r} & {\sf m} & {\sf a} & {\sf l} & {\sf i} & {\sf t} & 1 & d & 1 \\
\hline
8  & {\sf r} & {\sf m} & {\sf a} & {\sf l} & {\sf i} & {\sf t} & {\sf i} & 0 & 0 & 0 \\
\hline
9  & {\sf m} & {\sf a} & {\sf l} & {\sf i} & {\sf t} & {\sf i} & {\sf e} & 1 & d & 2 \\
\hline
10 & {\sf a} & {\sf l} & {\sf i} & {\sf t} & {\sf i} & {\sf e} & {\sf s} & 0 & 0 & 0 \\
\hline
11 & {\sf l} & {\sf i} & {\sf t} & {\sf i} & {\sf e} & {\sf s} & {\sf \_} & 0 & 0 & 0 \\
\hline
12 & {\sf i} & {\sf t} & {\sf i} & {\sf e} & {\sf s} & {\sf \_} & {\sf \_} & 1 & i & i \\
\hline
13 & {\sf t} & {\sf i} & {\sf e} & {\sf s} & {\sf \_} & {\sf \_} & {\sf \_} & 0 & 0 & 0 \\
\hline
\end{tabular}
\caption{Instances with morphological analysis classifications derived
from the word {\sf ab$|$norm$|$al$|$iti$|$es}. The three
classification fields belong 
to tasks {\sc m1}, {\sc m2}, and {\sc m3},
respectively. Denotations of the classification labels is as follows:
$0 =$ no morpheme boundary; $1 =$ morpheme boundary with {\sc m1}, and
stress-neutral affix with {\sc m3}; $2 =$ stress-affecting affix; $d =$
derivational boundary; $i =$ inflectional boundary; $s =$ stem
boundary.\label{morph-window-ex}}
\end{center}
\end{table}

As can be seen from Table~\ref{morph-window-ex}, a morpheme
boundary is assigned to the position at which a new morpheme
begins, regardless of the spelling changes that may have occurred in
the vicinity of that position. For example, the analysis displayed in
Table~\ref{morph-window-ex} states that the `surface' form {\sf iti} is a
stress-affecting affix, although its `deep' form is {\sf ity}. A second
characteristic of our representation of morpheme boundaries, is
that it is non-hierarchic. Although morpheme hierarchy may be
important in determining the part-of-speech of a word
\cite{allen-mitalk}, it is not necessary to have a full 
hierarchical analysis when the morphological analysis is used as input
to a text-to-speech system.

\section{Experiments}
\label{exps}

\subsection{Data Collection and Algorithmic Parameters}

The source for the morphological data used in our experiments is {\sc
celex} \cite{burnage-celex}, a large lexical data base of English,
Dutch, and German. We extracted from the English data base all
relevant information on wordforms relating to spelling and morphology,
and created a lexicon of 65,558 morphologically analysed words. This
lexicon was used to create instance bases for the {\sc
m1}, {\sc m2}, and {\sc m3} tasks, each containing 573,544 instances.

For completeness, the learning parameters of the five algorithms
described in section~\ref{intro}, viz. {\sc ib1}, {\sc ib1-ig}, {\sc
igtree}, {\sc c4.5}, and {\sc bp}, as used in our experiments, are the
following: (i) {\sc ib1} and {\sc ib1-ig} implement 1-nearest
neighbour matching; (ii) {\sc c4.5} uses the gain ratio criterion,
default pruning, and no subsetting of feature-values; (iii) {\sc bp}
uses a network with 294 input units (letters are locally coded), 50
hidden units, and 2, 3, or 5 output units (classes are locally coded),
a learning rate of 0.1, a momentum of 0.4, and an update tolerance of
0.2. {\sc igtree}'s functioning is not governed by parameters.

\subsection{Results}

\begin{figure}
\centerline{
        \epsfxsize=\textwidth
        \epsfbox{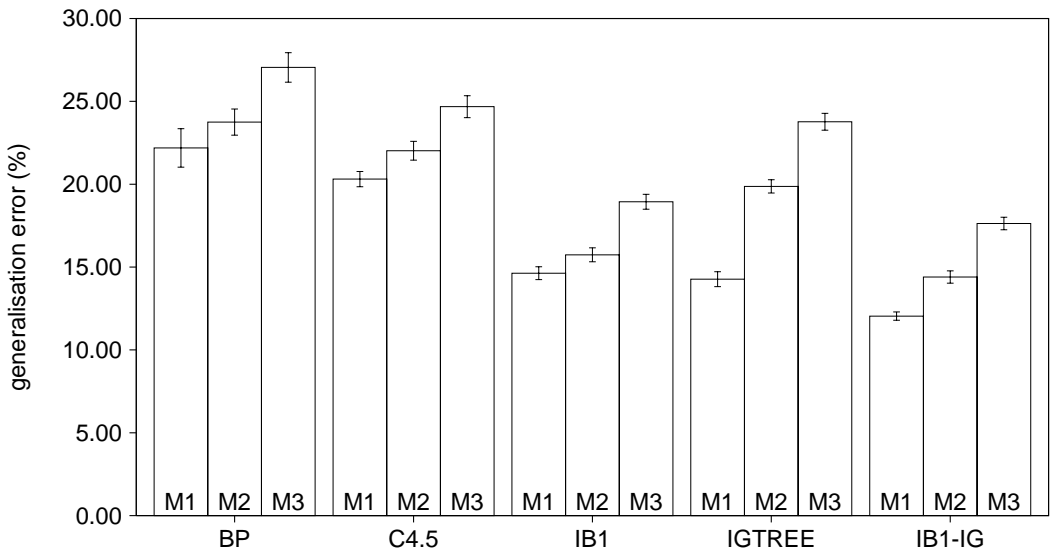} }
\center
\vspace*{-0.5cm}
\caption{Generalisation errors in terms of the
percentage of incorrectly classified test words, with standard
deviations (error bars) of five algorithms
applied to the three variations of the task of English
morphological analysis {\sc m1}, {\sc m2}, and {\sc m3}.} 
\label{morph-eng-gen}
\end{figure}

We applied the five algorithms to the three tasks, performing with
each algorithm and each task a 10-fold 
cross-validation experiment \cite{weiss-10foldcv}. We computed for
each 10-fold cross-validation experiment the average percentage of
incorrectly processed test words. A word is incorrectly processed when
{\sl one or more}\/ instance classifications associated with the instances
derived from the word are incorrect (i.e., when one or more of the
segmentations is incorrect). Figure~\ref{morph-eng-gen} displays
these generalisation errors. The algorithms
are ordered on their performance on task {\sc m1}.
 
The best performing algorithm on tasks {\sc m1}, {\sc m2}, and {\sc
m3} is {\sc ib1-ig}. Its performance is significantly better compared
to all other algorithms in all three tasks with $p<0.001$. On task {\sc
m1}, the algorithm performing second best to {\sc ib1-ig} (12.04\%
incorrectly processed test words) is {\sc igtree} (14.27\%)
(level of significance $t(19)=13.56, p<0.001$). On task {\sc m2}, the
second best algorithm is {\sc ib1} (15.74\%) ; {\sc ib1-ig} processes
14.40\% test words incorrectly; ($t(19)=7.64,
p<0.001$). On task {\sc 
m3}, {\sc ib1-ig} incorrectly processes 17.63\% of the test words,
again followed by {\sc ib1} with 18.94\% ($t(19)=6.95, p<0.001$).

\begin{figure}
\centerline{
        \epsfxsize=0.35\textwidth
        \epsfbox{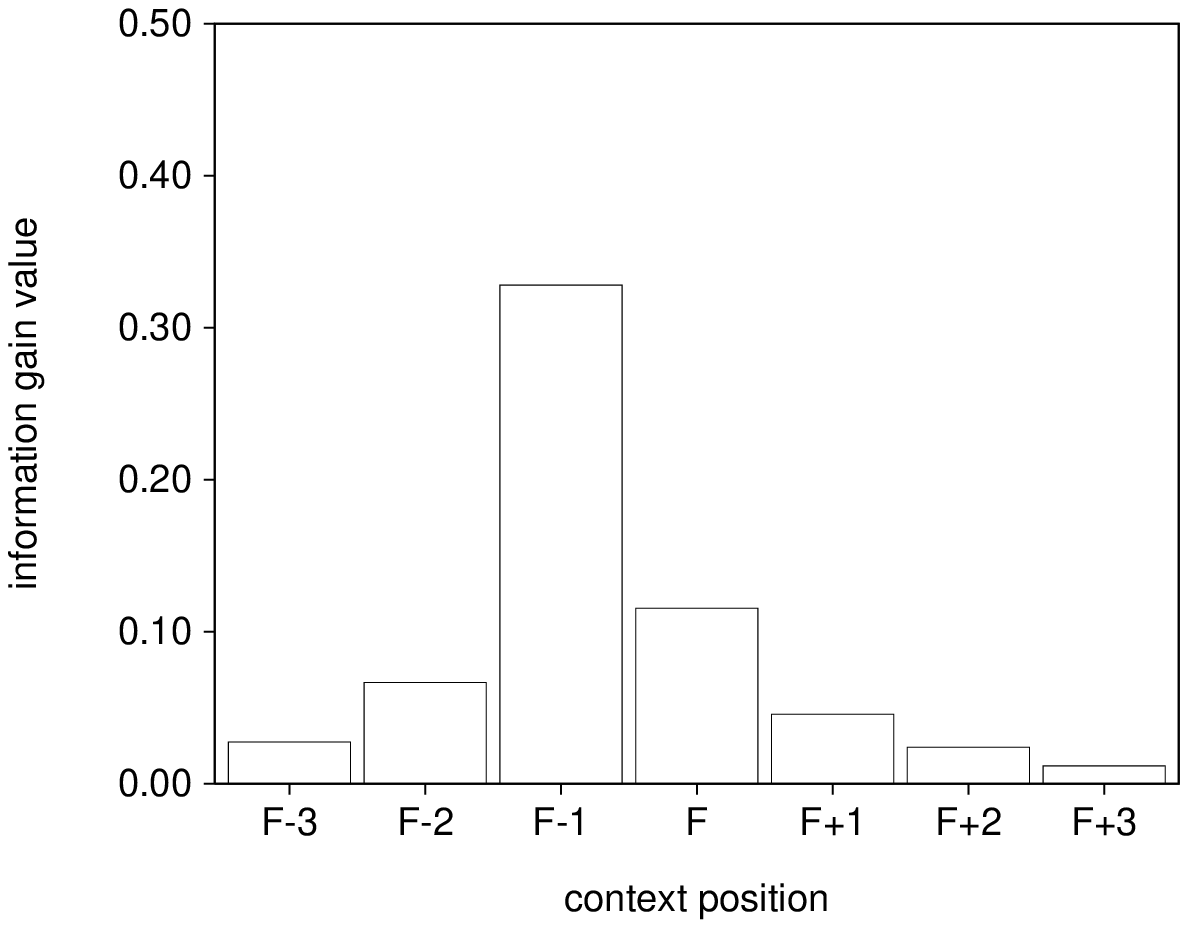}
        \epsfxsize=0.35\textwidth
        \epsfbox{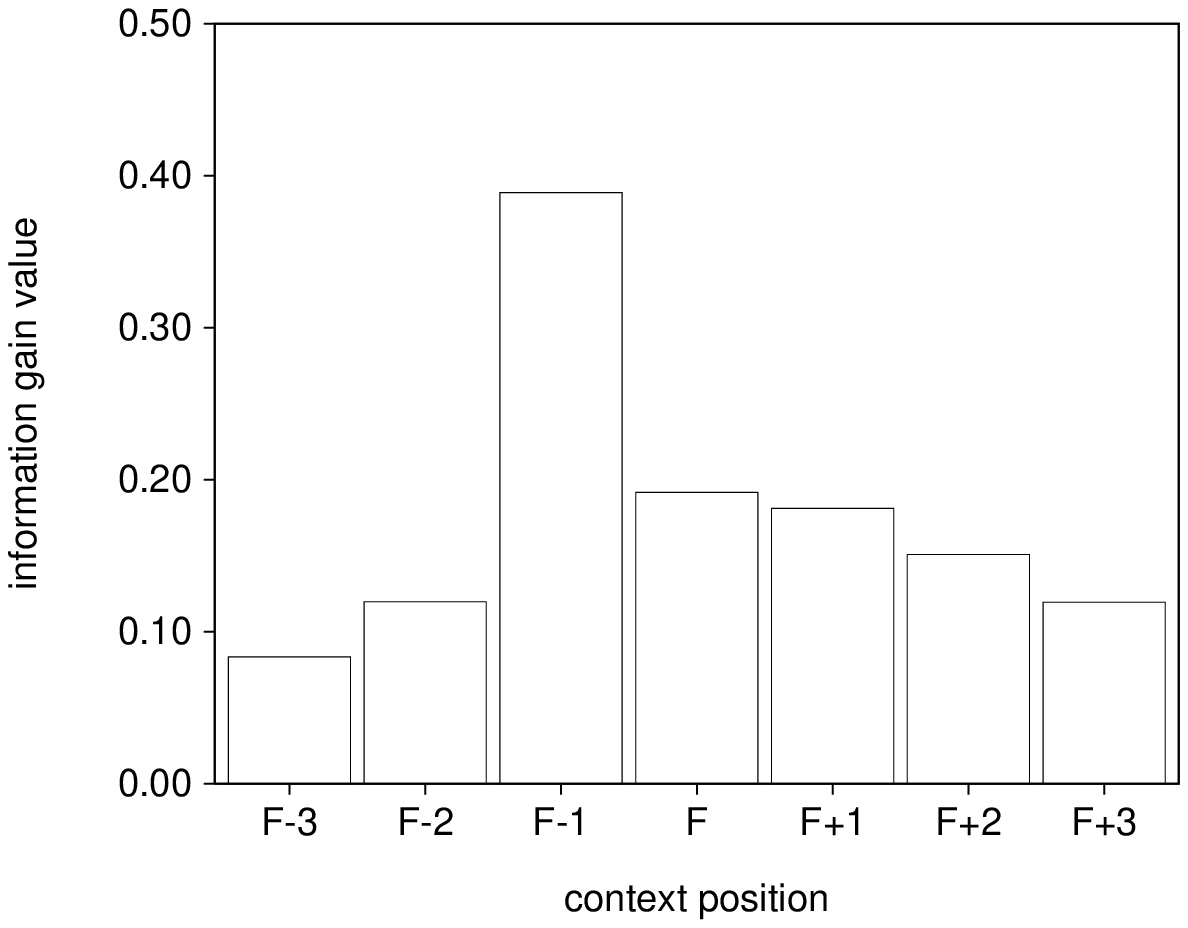}
        \epsfxsize=0.35\textwidth
        \epsfbox{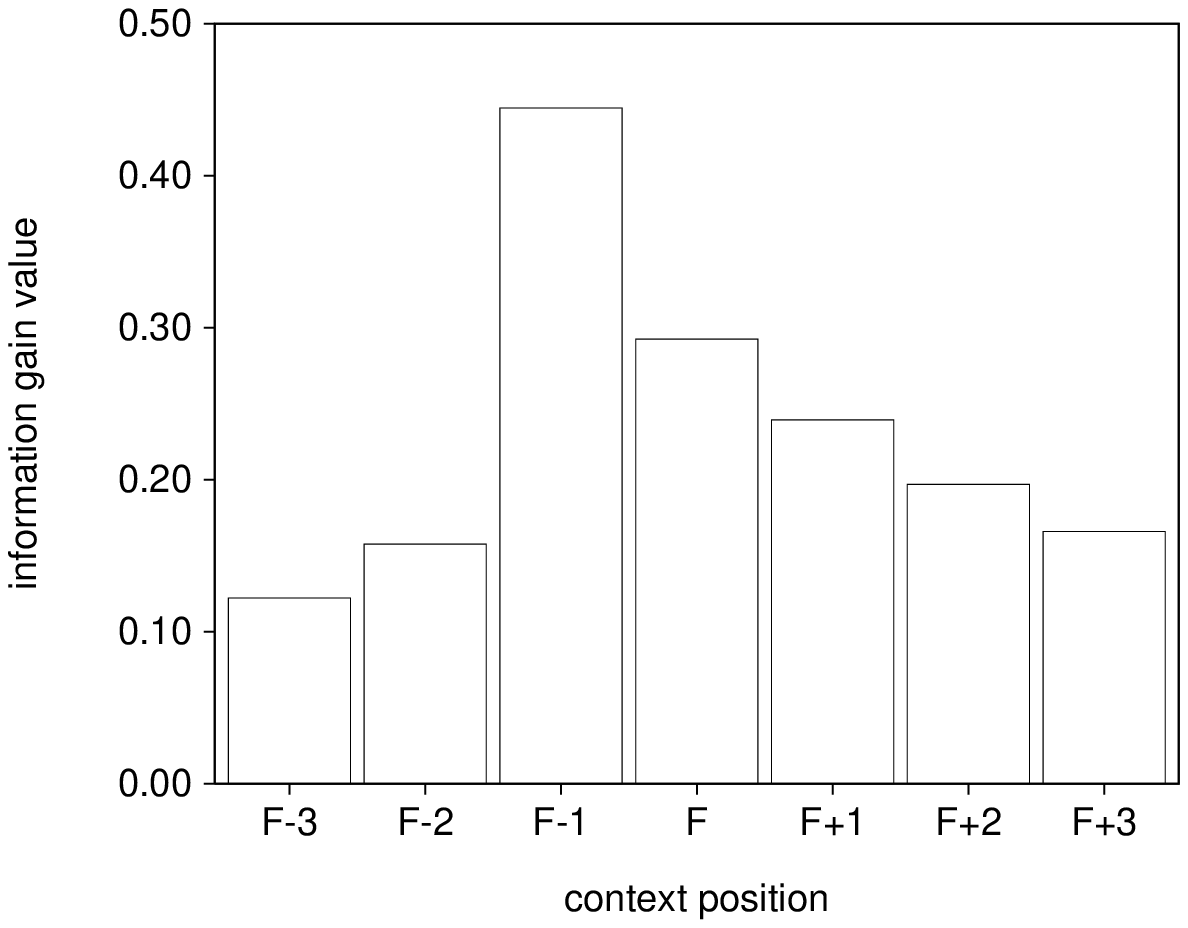}}
\center
\caption{Information-gain values of the features of tasks {\sc m1} (left),
{\sc m2} (middle), and {\sc m3} (right), computed over the full
instance bases.} 
\label{infogain}
\end{figure}

Interesting is the fact that {\sc igtree} performs well on {\sc m1},
but performs relatively bad on {\sc m2} and {\sc m3}. {\sc igtree} is
known to perform worse when the information gain of the input features
displays a low variance \cite{daelemans-air}, i.e., when there is
little difference between the `relative importance' of the input
features. This suggests
that the information-gain values of the features with tasks {\sc m2}
and {\sc m3} have less outspoken differences than with {\sc m1}, which
is indeed the case, as is displayed in Figure~\ref{infogain}.
For all three tasks, Figure~\ref{infogain} displays the fact the
letter immediately preceding the focus letter is the most important
one in the segmentation task.

A more general observations on the basis of the results displayed in
Figure~\ref{morph-eng-gen} is that tasks {\sc m1}, {\sc m2}, and {\sc
m3} are increasingly difficult to learn for all
algorithms. Distinguishing between more output classes with a finer
linguistic granularity obviously increases the difficulty of learning
the task.  The results in Figure~\ref{morph-eng-gen} also provide an
indication that the performance of the best algorithms is quite good,
considering (i) the test words are not seen by the algorithms during
training, and (ii) the test words are dictionary words, rather than
words from a written text corpus: they are on the average
morphologically more complex than words from a corpus. When the
generalisation performance is expressed in terms of incorrectly
classified instances, low error rates are obtained. For example,
trained on {\sc m1}, {\sc ib1-ig} classifies only 1.65\% of all test
instances incorrectly (1.97\% on {\sc m2}, and 2.46\% on {\sc m3}).

As an illustration, we provide some examples of segmentations
generated by {\sc ib1-ig} on the first partition of task {\sc
m1}. Most errors are related to (apparent) morphological ambiguities:
incorrect boundary insertions in {\sf ear$|$ly}, {\sf nav$|$y}, and
{\sf co$|$al$|$ed}, and missed boundaries in {\sf printable}, {\sf
upland}, and {\sf manslaught$|$er}. Some examples of correctly segmented
words that are morphologically complex are {\sf horse$|$whip}, {\sf
nut$|$ti$|$est}, {\sf steep$|$en}, {\sf veto$|$es}, and {\sf
dis$|$agree$|$able$|$ness}. 

\section{Conclusions}
\label{conclusions}

We have demonstrated the applicability of an inductive
machine-learning approach to morphological analysis, by reformulating
the 
problem as a segmentation task in which letter sequences are
classified as marking different types of morpheme boundaries. The
generalisation performance of inductive-learning algorithms to the
task is good.

An interesting result is that within the class of inductive learning
algorithms, generalization accuracy correlates with the degree of
eagerness of the inductive algorithm used; best results are obtained
with memory-based learning ({\sc ib1-ig}), a lazy learning algorithm
retaining full memory of all training instances with a
classification-task-related feature-weighting similarity function. The
methods abstracting most from the instances perform worst. This
corroborates our hypothesis that because of the intricate interaction
of regularities, subregularities and exceptions present in this task
as well as in most other linguistic problems we studied, lazy learning
methods are superior to eager learning methods.

In comparison with the traditional approach, in which morphological
analysis is performed by a system containing several components, the
inductive learning approach applied to a reformulation of the problem
as a classification task of the segmentation type, has a number of
advantages:

\begin{itemize}
\item
it presupposes no more linguistic knowledge than explicitly present in
the corpus used for training, i.e., it avoids a knowledge-acquisition
bottleneck;  
\item
it is language-independent, as it functions on any morphologically
analysed corpus in any language;
\item
learning is automatic and fast;
\item
processing is deterministic, non-recurrent (i.e., it does not retry
analysis generation) and fast, and is only linearly related to the
length or morphological complexity of words.
\end{itemize}
\ \\[-0.7cm]

Nevertheless, it also displays two disadvantages:

\begin{itemize}
\item
it produces an analysis that lacks hierarchy of morphemes;
\item
it does not recover the `deep' form of morphemes.
\end{itemize}

Future work on inductive learning of morphological analysis should
include a thorough performance comparison with existing traditional
systems for morphological analysis, based on linguistic theory and
heuristics such as {\sc decomp} \cite{allen-mitalk} as well as with
probabilistic systems \cite{heemskerk-morpa}. Secondly, we aim at
integrating trained models of morphological analysis into
larger systems, to investigate whether the enrichment of spelling
input with morpheme boundary information improves the
generalisation performance of other learning systems trained on, e.g.,
stress assignment, grapheme-phoneme conversion, and part-of-speech
prediction of unknown words.

\newcommand{\noopsort}[1]{} \newcommand{\printfirst}[2]{#1}
  \newcommand{\singleletter}[1]{#1} \newcommand{\switchargs}[2]{#2#1}

\end{document}